\pdfoutput=1
\documentclass[11pt]{article}

\usepackage{acl}

\usepackage{times}
\usepackage{latexsym}

\usepackage[T1]{fontenc}
\usepackage[utf8]{inputenc}

\usepackage{microtype}

\usepackage{tabularx}
\usepackage{graphicx}
\usepackage{array}
\usepackage{float}
\restylefloat{table}
\usepackage{adjustbox}
\usepackage{array, multirow}
\usepackage{comment}
\usepackage{xcolor}
\usepackage{amsmath}
\usepackage{bm}

\usepackage[english]{babel}
\usepackage{blindtext}
\usepackage{bbm}

\usepackage{color}
\usepackage{tcolorbox}
\usepackage{CJK}
\usepackage{adjustbox}
\usepackage{booktabs}
\tcbset{width=0.4\textwidth,boxrule=0pt,colback=red,arc=0pt,auto outer arc,left=0pt,right=0pt,boxsep=5pt}
\usepackage{cleveref}
\usepackage{amsmath}
\newcolumntype{P}[1]{>{\centering\arraybackslash}p{#1}}
\usepackage{pgfplots}
\pgfplotsset{compat=1.17}
\usepackage{stackengine}

\usepackage{physics}
\usepgfplotslibrary{groupplots}
\usepackage{amsbsy}
\title{Transferring Knowledge via Neighborhood-Aware Optimal Transport \\ for Low-Resource Hate Speech Detection}

\author{Tulika Bose \quad Irina Illina  \quad Dominique Fohr  \\
  Universite de Lorraine, CNRS, Inria, LORIA, F-54000 Nancy, France\\
  {\tt \{tulika.bose, illina, dominique.fohr\}@loria.fr}\\ 
}

\begin{document}
\maketitle
\begin{abstract}

\emph{{\bf Warning: } this paper contains content that may be offensive and distressing.}

The concerning rise of hateful content on online platforms has increased the
attention towards automatic hate speech detection, commonly formulated as a supervised classification task. 
State-of-the-art deep learning-based approaches usually require a substantial amount of labeled resources for training. However, annotating hate speech resources is expensive, time-consuming, and often harmful to the annotators. This creates a pressing need to transfer knowledge from the existing labeled resources to low-resource hate speech corpora with the goal of improving system performance. For this, neighborhood-based frameworks have been shown to be effective. However, they have limited flexibility.  
In our paper, we propose a novel training strategy that allows flexible modeling of the relative proximity of neighbors retrieved from a resource-rich corpus to learn the amount of transfer. 
In particular, we incorporate neighborhood information with Optimal Transport, which permits exploiting the geometry of the data embedding space.
By aligning the joint embedding and label distributions of neighbors, we demonstrate substantial improvements over strong baselines, 
in low-resource scenarios, on different publicly available hate speech corpora.

\end{abstract}

\section{Introduction}

With the alarming spread of Hate Speech (HS) in social media, 
Natural language Processing techniques have been used to develop automatic HS detection systems, typically to aid manual content moderation. 
Although deep learning-based approaches \citep{mozafari2019bert, 10.1145/3041021.3054223} have become state-of-the-art in this task, their performance depends on the size of the labeled resources available for training \citep{lee-etal-2018-comparative, alwosheel2018your}. 

Annotating a large corpus for HS
is considerably time-consuming, expensive, and harmful to human annotators \citep{schmidt-wiegand-2017-survey, Malmasi2018ChallengesID, Poletto2019AnnotatingHS, sarwar-etal-2022-neighborhood}. Moreover, models trained on existing labeled HS corpora have shown poor generalization when evaluated on new HS content \citep{ yin2021generalisable, 10.1145/3331184.3331262, swamy-etal-2019-studying, karan-snajder-2018-cross}. This is due to the differences across these corpora, such as sampling strategies 
\citep{wiegand-etal-2019-detection}, varied topics of discussion \citep{app10124180, 10.1145/2124295.2124376}, varied vocabularies, and different victims of hate. Thus, to address these challenges, here we aim to devise a strategy that can effectively transfer knowledge from a resource-rich source corpus with a higher amount of annotated content to a low-resource target corpus with fewer labeled instances.

One popular way to address this is transfer learning. For instance, 
\citet{mozafari2019bert} 
fine-tune a large-scale pre-trained language model, BERT \citep{devlin-etal-2019-bert}, on the limited training examples in HS corpora. Further, a sequential transfer, following \citet{Garg2020TANDATA}, can be performed where a pre-trained model is first fine-tuned on a resource-rich source corpus and subsequently fine-tuned on the low-resource target corpus. Since this may risk forgetting knowledge from the source, the source and target corpora can be mixed for training \citep{shnarch-etal-2018-will}. Besides, to learn target-specific patterns without forgetting the source knowledge, \citet{Meftah2021NeuralSD} augment pre-trained neurons from the source model with randomly initialized units for transferring knowledge to low-resource domains. 

Recently, \citet{sarwar-etal-2022-neighborhood} argue that traditional transfer learning strategies are not systematic. Therefore, they model the relationship between a source and a target corpus
with a neighborhood framework and show its effectiveness in transfer learning for content flagging. They model the interaction between a query instance from the target and its neighbors retrieved from the source. This interaction is modeled based on their label agreement -- whether the query and its neighbors have the same labels -- while using a fixed neighborhood size. However, different neighbors may have varying levels of proximity to the queried instance based on their pair-wise cosine similarities in a sentence embedding space. Therefore, intuitively, the neighbors should also be weighted according to these similarity scores. 

We hypothesize that simultaneously modeling the pair-wise distances between instances from the low-resource target and their respective neighbors from the resource-rich source, along with 
their label distributions should result in a more flexible and effective transfer. With this aim, we propose a novel training strategy where the model learns to assign varying importance to
the neighbors corresponding to different target instances by optimizing the amount of pair-wise transfer. This transfer is learned without changing the underlying model architecture.
Such optimization can be efficiently performed using \textit{Optimal Transport} (OT) \citep{MAL-073, Villani2008OptimalTO, kantorovich2006translocation} 
%monge1781memoire} 
due to its ability to find correspondences between instances while exploiting the underlying geometry of the embedding space. 
Our contributions are summarised as follows:
\begin{itemize}
    \item We address HS detection in low-resource scenarios with a flexible and systematic transfer learning strategy.  
    \item We propose novel incorporation of neighborhood information with joint distribution Optimal Transport. This enables learning of the amount of transfer 
    between pairs of source and target instances considering both (i) the similarity scores of the neighbors and (ii) their associated labels.
    To the best of our knowledge, this is the first work that introduces Optimal Transport for HS detection.
    \item We demonstrate the effectiveness of our approach through considerable improvements over strong baselines, along with quantitative and qualitative analysis on different HS corpora from varied platforms.
\end{itemize}

\section{Related Works}

\subsection{Hate Speech Detection}
Deep Neural Networks, especially the transformer-based models,
such as the pre-trained BERT, have
dominated the field of HS detection in the past few years \citep{Alatawi2021DetectionOH, DSa2020BERTAF, glavas-etal-2020-xhate, mozafari2019bert}. 

\citet{wiegand-etal-2019-detection, 10.1145/3331184.3331262} raise concerns about data bias present in most HS corpora, which results in overestimated within-corpus performance. They, therefore, recommend cross-corpus evaluations as more realistic settings.  \citet{bigoulaeva-etal-2021-cross, bose-etal-2021-generalisability, PAMUNGKAS2021102544} perform such cross-corpus evaluations
%zero-shot transfer learning setups with 
in this task
with no access to labeled instances from the target. However, \citet{yin2021generalisable, wiegand-etal-2019-detection} report fluctuating or degraded performance across corpora. As pointed out by \citet{sarwar-etal-2022-neighborhood}, in real-life scenarios, most online platforms could invest in obtaining at least some labeled training instances for deploying an HS detection system. Thus, we study a more realistic setting where a limited amount of labeled content is available in the target corpus.

\subsection{Neighborhood Framework}
\label{related_neigh}

$k$-Nearest Neighbors (${k}$NN)-based approaches have been successfully used in the literature for an array of tasks such as language modeling \citep{Khandelwal2020Generalization}, question answering \citep{kassner-schutze-2020-bert}, dialogue generation \citep{10.1162/tacl_a_00356}, etc.
Besides, ${k}$NN classifiers have been used for HS detection \citep{prasetyo2022hate, Briliani2019HateSD}, which typically predict the class of an input instance through a simple majority voting using its neighbors in the training data. 

Recently, \citet{sarwar-etal-2022-neighborhood} propose a neighborhood framework $k$NN$^{+}$ for transfer learning in cross-lingual low-resource settings.
They show that a simple \textit{k}NN classifier is prone to prediction errors as the neighbors may have similar meanings, but opposite labels. 
They, instead, model the interactions between the target corpus instances, treated as queries, and their nearest neighbors retrieved from the source. This neighborhood interaction is modeled based on whether a query and its neighbors have the same or different labels. In their best performing framework (in cross-lingual setting) of Cross-Encoder $k$NN$^{+}$, \citet{sarwar-etal-2022-neighborhood} obtain representations of concatenated query-neighbor pairs to learn such neighborhood interactions.

However, \citet{sarwar-etal-2022-neighborhood} do not 
consider \textit{the varying levels of the proximity of different neighbors to the query.} Besides, a mini-batch in their framework comprises a query and all its neighbors. 
For fine-tuning large language models like BERT, the batch size needs to be kept small due to resource constraints. This could limit the neighborhood size in their framework.
This is different from 
our approach, where the neighborhood size is scalable.

\subsection{Optimal Transport}
Optimal Transport (OT) has become increasingly popular in diverse NLP applications, as it allows comparing probability distributions in a geometrically sound manner. These include  
machine translation \citep{xu-etal-2021-vocabulary}, 
interpretable semantic similarity \citep{lee-etal-2022-toward}, 
rationalizing text matching \citep{swanson-etal-2020-rationalizing}, etc. Moreover, OT has been successfully used for domain adaptation in audio, images, and text \citep{olvera2021improving, damodaran2018deepjdot, chen2020graph}.
In this work, we perform novel incorporation of nearest neighborhood information with OT. 
Besides, to the best of our knowledge, this is the first work that introduces OT to the HS detection task.

\section{Proposed Approach}

Our problem setting involves a low-resource target corpus $X^{t}$ with a limited amount of labeled training data $(X^t_{train}, Y^t_{train}) = \{x_i^t, y_i^t\}_{i=1}^{n_t}$ and a resource-rich source corpus $X^{s}$ from a different distribution with a large number of annotated data $(X^{s}_{train}, Y^{s}_{train}) = \{x_i^s,y_i^s\}_{i=1}^{n_s}$. Given such a setting, we hypothesize that transferring knowledge from the nearest neighbors in the source  should improve the performance on the insufficiently labeled target.  
Furthermore, to provide additional control to the model, we propose a systematic transfer. With this transfer mechanism, a model can 
\textit{learn} different weights assigned to the neighbors in $X^{s}_{train}$ based on their proximity to the 
instances in $X^{t}_{train}$ simultaneously in a sentence embedding space and the label space. For this, we incorporate neighborhood information with Optimal Transport (OT), as OT can learn correspondences between instances from $X^{s}_{train}$ and $X^{t}_{train}$ by exploiting the underlying embedding space geometry.

\subsection{Joint Distribution Optimal Transport}
 
In this work, we use the joint distribution optimal transport (JDOT) framework \citep{10.5555/3294996.3295130}
following the works of \citet{damodaran2018deepjdot, fatras2021unbalanced}, proposed for unsupervised domain adaptation in deep embedding spaces. The framework aligns the joint distribution $P(Z, Y)$
of the source and the target domains, where $Z$ is the embedding space through a mapping function $g(.)$, and $Y$ is the label space.
For a discrete setting, let $\mu_s = \sum_i^{n_s}a_i\,\delta_{g(x_i^s),y_i^s}$ and $\mu_t = \sum_i^{n_t}b_i\,\delta_{g(x_i^t),y_i^t}$ be two empirical distributions on the product space of $Z \times Y$. Here 
$\delta_{g(x_i),y_i}$ is the Dirac function at the position $(g(x_i),y_i)$, and $a_i$, $b_i$ are uniform probability weights, i.e. $\sum_i^{n_s}a_i = \sum_i^{n_t}b_i = 1$. 

The `balanced' OT problem ($OT_b$), as defined by \citet{kantorovich2006translocation}, seeks for a transport plan $\gamma$ in the space of the joint probability distribution $\Pi(\mu_s,\mu_t)$, with marginals $\mu_s$ and $\mu_t$, that minimizes the cost of transport from $\mu_s$ to $\mu_t$, as:

% \begin{equation}\label{eq0}
% \resizebox{0.7\columnwidth}{!}{%

% %\begin{math}
% %\[
% %\begin{math}
% \begin{aligned}
%     {OT_b}(\mu_s,\mu_t)  &= \underset{\gamma \in \Pi(\mu_s,\mu_t)}{\mathrm{min}} \quad  \underset{i,j}{\sum} \gamma_{i,j}c_{i,j} \\
%     & s.t. \quad  \gamma{\vb 1}_{n_t} = \mu_s, \gamma^T{\vb 1}_{n_s} = \mu_t
    
% \end{aligned}
% %\end{math}
% %\]
% %\end{math}

% %\]
% }
% \end{equation}
\begin{equation} \label{eq0}
\resizebox{0.7\columnwidth}{!}{%
$\begin{aligned}
{OT_b}(\mu_s,\mu_t)  &= \underset{\gamma \in \Pi(\mu_s,\mu_t)}{\mathrm{min}} \quad  \underset{i,j}{\sum} \gamma_{i,j}c_{i,j} \\
    & s.t. \quad  \gamma{\vb 1}_{n_t} = \mu_s, \gamma^T{\vb 1}_{n_s} = \mu_t
\end{aligned}$
}
\end{equation}
Here $c_{i,j}$ is an entry in a cost matrix $C \in R^{n_s \times n_t}$, representing the pair-wise cost (see Section \ref{neigh-OT}), and $\vb{1}_n$ is a vector of ones with dimension $n$. Each entry $\gamma_{i,j}$ indicates the amount of transfer from location $i$ in the source to $j$ in the target. 

The constraint on $\gamma$ requires that all mass from $\mu_s$ is transported to $\mu_t$. However, this can be alleviated through relaxation, leading to the `unbalanced' OT ($OT_u$) \citep{M2AN_2003__37_5_851_0}, as:

% \begin{equation} \label{eq1}
% \resizebox{\columnwidth}{!}{%
% $\begin{aligned}
% {OT_u}(\mu_s,\mu_t)  &= \underset{\gamma \in \Pi(\mu_s,\mu_t)}{\mathrm{min}} \quad  \underset{i,j}{\sum} \gamma_{i,j}c_{i,j} + \Lambda\,;\\
%     & \text{where} \quad \Lambda = \epsilon \: \Omega(\gamma)
%      + \lambda\left(\text{KL}(\gamma\vb{1}_{n_t},\mu_s) + \text{KL}(\gamma^T\vb{1}_{n_s}, \mu_t)\right) \\
%     & s.t. \quad \gamma \geq 0
% \end{aligned}$
% }
% \end{equation}

\begin{equation}\label{eq1}
\resizebox{\columnwidth}{!}{%
$\begin{aligned}
{OT_u}(\mu_s,\mu_t)  &= \underset{\gamma \in \Pi(\mu_s,\mu_t)}{\mathrm{min}} \quad  \underset{i,j}{\sum} \gamma_{i,j}c_{i,j} + \Lambda\,;\\
    & \text{where} \quad \Lambda = \epsilon \: \Omega(\gamma)
     + \lambda\left(\text{KL}(\gamma\vb{1}_{n_t},\mu_s) + \text{KL}(\gamma^T\vb{1}_{n_s}, \mu_t)\right) \\
    & s.t. \quad \gamma \geq 0
\end{aligned}$
}
\end{equation}
$\text{KL}$ is the Kullback-Leibler divergence that allows the relaxation of the marginal constraint on $\gamma$. $\lambda$ is the marginal relaxation coefficient. $\Omega(\gamma) = \sum_{i,j}\gamma_{i,j}log(\gamma_{i,j})$ corresponds to the entropic regularization term, which allows fast computation of the OT distances \citep{NIPS2013_af21d0c9}. 
$\epsilon$ is the entropy coefficient.

% \begin{figure*}[t]
% \begin{minipage}{\textwidth}
% \centering 
% \includegraphics[width=1.05\textwidth, height=7.1cm]{OT_NN.pdf} \qquad
% \captionof{figure}{Illustration of the training strategy in OT$^{NN}$. Even though the BERT encoder $g$ and the classifier $f$ are shared by both corpora, they are illustrated
% twice for better clarity by representing the two corpora separately. The presented softmax values obtained from $f$ are simply examples provided for illustration. The figure is inspired by \citet{damodaran2018deepjdot}.} 
% \label{OT_NN_fig}
% \end{minipage}
% \end{figure*}

For models with a high-dimensional embedding space like ours,
\citet{fatras2021unbalanced} propose to make the computation of OT losses scalable 
using the mini-batch OT. 
Thus, for every mini-batch, we sample an equal number of instances, given by the batch size $m$, from $X^{s}_{train}$ and $X^{t}_{train}$, which makes $C \in R^{m \times m}$ and $\gamma$ square matrices. 
As discussed by \citet{fatras2021unbalanced}, since the transport plan at the mini-batch level is much less sparse, it may result in undesired pairings between instances if computed by Equation \ref{eq0}. %\citep{fatras2021unbalanced}. 
To counteract this effect, we rely on the more robust version of OT as formulated in Equation \ref{eq1}.
Thus, we adopt the \textit{joint distribution entropy regularized unbalanced mini-batch OT} for our framework, henceforth simply referred to as OT. Note that this framework does not modify the underlying model architecture used for 
classification, but only introduces a new training strategy.

\subsection{Neighborhood-aware OT (OT$^{\boldsymbol{NN}}$)}
\label{neigh-OT}

In the above joint distribution framework, the cost matrix $C$ is expressed as the weighted combination of the costs in the embedding and the label spaces: %given by Equation \ref{eq2}.
\begin{equation}\label{eq2}
\resizebox{\columnwidth}{!}{%
$\textit{c}_{i,j}(g(x_i^s),y_i^s; g(x_j^t),y_j^t) = \alpha\:d(g(x_i^s),g(x_j^t))
    +\beta\:L(y_i^s,y_j^t)$
}
\end{equation}
$d(.,.)$ denotes the \textit{embedding distance} (ED), which is a squared $l_2$ distance between the corresponding embeddings. $L(.,.)$ is \textit{label-consistency loss} (LC), which is a cross-entropy loss that enforces a match between the label of the $i^{th}$ source instance and that of the $j^{th}$ target instance.
%respectively from $X^{s}_{train}$ and $X^{t}_{train}$. 
$\alpha$ and $\beta$ are scalar values. Minimizing the cost in Equation \ref{eq2} results in aligning instances from the source and the target that simultaneously share similar representations and common labels. 

We adapt $C$ to account for $k$ nearest neighbors of the target instances in $X^{t}_{train}$ from the source $X^{s}_{train}$. Since BERT is not optimal for semantic similarity search \citep{reimers-gurevych-2019-sentence}, we extract the neighbors using the Sentence-BERT (SBERT) model \citep{reimers-gurevych-2019-sentence}. SBERT provides sentence embeddings that can be easily compared using cosine similarity.  We hypothesize that allowing transfers to occur only from the corresponding neighbors in the source to the target should result in more effective learning.

For this, we explicitly assign the value $\mathrm{max}(C)$ to $c_{i,j}$ in $C$ whenever the $i^{th}$ source and $j^{th}$ target instances are not neighbors, considering the nearest neighborhood space of $k$ neighbors. Besides, we use the SBERT distances as the embedding distance in Equation \ref{eq2}. This distance, in addition to the label consistency term, ensures that $\gamma$ is learned to allow a higher amount of transfer from neighbors in $X^{s}_{train}$ that are simultaneously (i) closer in the SBERT space and (ii) share the same label with an instance in $X^{t}_{train}$, compared to the neighbors that are further away and\slash or have opposite labels.

\textit{Note that even though we use a neighborhood size of $k$, the target instances do not attend equally to all of their $k$ neighbors.} This is because if the distance between a target instance $x_j^{t}$ and its top $n^{th}$ neighbor ($x_i^{s}$) from the source, within the neighborhood size of $k$ (i.e.  $n$ < $k$) is comparatively large, their corresponding $(i,j)\mbox{-}th$ entry in $C$ would have a larger value. This would comparatively reduce the transfer \textit{even if they share common labels}. Thus, for a neighbor with the same label as the target instance, the higher its SBERT distance from the target instance, the lower the amount of transfer.
This results in more flexibility where the model can learn 
from the relevant neighbors corresponding to every target instance.

In addition to the OT loss from Equation \ref{eq1}, we introduce the cross-entropy losses for the training instances from both $X^{t}_{train}$ and $X^{s}_{train}$ in the final loss function, as required by our classification task. Our final loss function is given by Equation \ref{eq3}. Here $g(.)$ encodes a given input using the pre-trained BERT encoder to the BERT embedding space 
by extracting the fine-tuned [CLS] token representation of the last hidden layer. $f(.)$ denotes the classifier, which is one fully connected layer. 
%that maps the representation into the classes.
$\theta_s$ and $\theta_t$ are the weights assigned to the source and the target cross-entropy losses, respectively. 
\begin{equation}\label{eq3}
\resizebox{0.9\columnwidth}{!}{%
$\begin{split}
    \mathrm{OT}^{NN}  =  \underset{\gamma,f,g}{\mathrm{min}}  \quad &%\left(
    \theta_s \frac{1}{m} \underset{i}{\sum} L_{s}\left(y_i^{s},f(g(x_i^{s}))\right) + \underset{i,j}{\sum}\gamma_{i,j}c_{i,j}
  %    \theta_s \frac{1}{m} \underset{i}{\sum}  L_{s}(y_i^{s},f(g(x_i^{s}))) +
      \\
     & + \Lambda +  \theta_t \frac{1}{m} \underset{j}{\sum} L_{t}\left(y_j^{t}, f(g(x_j^{t}))\right)
%     + \theta_t \frac{1}{m} \underset{j}{\sum} L_{t}(y_j^{t}, f(g(x_j^{t}))); \\
%   & \text{where} \quad \Lambda = \epsilon\,\Omega(\gamma) + \lambda\,\left(\text{KL}(\gamma\vb{1}_m,\mu_s) + \text{KL}(\gamma^T\vb{1}_m, \mu_t)\right)
\end{split}$
}
\end{equation}

\paragraph{Solving the optimization problem:}
Following \citet{damodaran2018deepjdot}, we adopt a two-step procedure to solve the above optimization problem at the mini-batch level. We first compute the optimal $\gamma$ by fixing the model parameters of $f$ and $g$.
\begin{equation}\label{eq4}
\resizebox{\columnwidth}{!}{%
$\underset{\gamma}{\mathrm{min}}
    \underset{i,j}{\sum}\gamma_{i,j}\left(\alpha\, d({g_{sbert}}(x_i^{s}),{g_{sbert}}(x_j^{t}))+\beta\, L(y_i^{s},y_j^{t})\right) + \Lambda $
  }
\end{equation}

We use the SBERT embeddings through the mapping function $g_{sbert}(.)$ here instead of the learned BERT embeddings to compute the ED loss. This is done so that the $\gamma$ is updated based on the semantic proximity in the SBERT space. $y_i^{s}$ and $y_j^{t}$ are the ground truth labels for the instances $x_i^{s}$ and $x_j^{t}$ from $X^{s}_{train}$ and $X^{t}_{train}$, respectively.
In the next step, the model parameters of $f$ and $g$ are learned while fixing $\gamma$ obtained from Equation \ref{eq4}, denoted as  $\hat{\gamma}$.
\begin{equation}\label{eq5}
\resizebox{\columnwidth}{!}{%
$\begin{aligned}
\underset{f,g}{\mathrm{min}} \quad 
     & \underset{i,j}{\sum}\hat{\gamma}_{i,j}\left(\alpha\: d(g(x_i^{s}),g(x_j^{t}))+\beta\:L(f(g(x_i^{s})),y_j^{t})\right) \\
    & + \theta_s \frac{1}{m} \underset{i}{\sum} L_{s}\left(y_i^{s},f(g(x_i^{s}))\right) +  \theta_t \frac{1}{m} \underset{j}{\sum} L_{t}\left(y_j^{t}, f(g(x_j^{t}))\right)
\end{aligned}$
}
\end{equation}
The first part of Equation \ref{eq5} allows the model to learn from the instances in $X^{s}_{train}$ that are consistent in terms of both the embedding space (ED loss) and the label space (LC loss) with the instances in $X^{t}_{train}$. Here we use $g(.)$, instead of $g_{sbert}(.)$, to compute ED so that $g$ learns from the SBERT space
through $\hat{\gamma}$. For the LC loss, we use the predicted labels for $x_i^{s}$ from the source and the actual labels $y_{j}^{t}$ corresponding to $x_j^{t}$ from the target. This is done to update the model parameters of $f$ and $g$ based on the target labels and bring source instances that have common labels closer to the target instances. We have provided an illustration of the training strategy of OT$^{NN}$ in Figure \ref{OT_NN_fig} of Appendix \ref{OT-fig}.

We propose different variants of OT$^{NN}$:
\paragraph{$\boldsymbol{\mathrm{OT}}^{\boldsymbol{NN}}$:} In this variant, we do not use the source cross-entropy loss term in Equation \ref{eq3}, thus effectively having $\theta_s = 0$.

\paragraph{$\bm{\mathrm{OT}^{NN}}_{\bm{\mathrm{pre\mbox{-}select}}}$:} Prior to the training, we pre-select the $k$ nearest neighbors from $X^s_{train}$ corresponding to every instance in $X^t_{train}$, instead of training with all the source instances.
Here also $\theta_s = 0$.

\paragraph{$\bm{\mathrm{OT}}^{\bm{NN}}$ + sloss:} This is OT$^{NN}$ with source cross-entropy loss (sloss), thus having $\theta_s = 1$.

\paragraph{$\bm{\mathrm{OT}^{NN}}_{\bm{\mathrm{pre\mbox{-}select}}}$ + sloss:} This is similar to the second variant, with $\theta_s = 1$. Here, sloss is computed only on the pre-selected source instances.

\section{Experimental Settings}

\subsection{Corpus Description}
\label{corp}
We perform experiments with three standard HS corpora, namely, \textit{Waseem} \citep{waseem}, \textit{Vidgen} \citep{vidgen-etal-2021-learning}, and \textit{Ethos} \citep{mollas_ethos_2022}, as they are collected using different sampling strategies across varied platforms. Following  
\citet{wiegand-etal-2019-detection, swamy-etal-2019-studying}, we use the labels of \textit{hate} and \textit{non-hate}, where the former involves all forms of hate.

\textit{Waseem} is a Twitter corpus comprising hate against women and ethnic minorities. 
We obtain 10.9K tweets in total from the tweet IDs, of which  26.8\% instances belong to the \textit{hate} class. \textit{Vidgen} is collected using a human-and-model-in-the-loop process aimed at making the corpus robust. It covers hate against diverse social groups, like blacks, women, muslims, immigrants, etc. with a total of 41144 instances, of which 53.9\% is labeled as \textit{hate}. \textit{Ethos} comprises 998 instances from YouTube and Reddit, of which 43.4\% are \textit{hate} instances. Even with fewer instances, it is made diverse with an active learning-based sampling strategy, ensuring a balance with respect to different hateful aspects. See Appendix \ref{corp_det} for further details on the corpora.

For our experiments, we create two different versions of every corpus depending on its use as the source or the target, as presented in Table \ref{dataset-prop}.

\begin{table}[!h]
\scriptsize
%\centering
\begin{center}
\begin{tabularx}{0.85\columnwidth}{ p{1.9cm} | p{0.85cm} | P{1.5cm} | p{0.85cm} }
\hline
{\bf Corpus}  & \multicolumn{3}{P{3.2cm}} {\bf Number of comments}   \\ 
\hline
\multicolumn{4}{c}{ \textbf{Source setting}} \\ \hline
&\multicolumn{3}{c}{\bf Train}  \\ \hline

\textit{Waseem}$_\mathrm{src}$
& \multicolumn{3}{c}{8720}  \\ 
\textit{Vidgen}$_\mathrm{src}$
& \multicolumn{3}{c}{32924}  \\ 
\textit{Ethos}$_\mathrm{src}$
& \multicolumn{3}{c}{998} \\
\hline 
\multicolumn{4}{c} {\bf Target setting} \\
\hline
&\bf Train & \bf Validation & \bf Test  \\ \hline
\textit{Waseem}$_\mathrm{tar}$
& 400 & 100 & 1090 \\ 
\textit{Vidgen}$_\mathrm{tar}$
& 400 & 100 & 4120  \\ 
\textit{Ethos}$_\mathrm{tar}$
& 400 & 100 & 200 \\
\hline
\end{tabularx}
\end{center}
\caption{\label{dataset-prop} Corpus statistics.}.
\end{table}

\paragraph{Source setting:} In the absence of available standard splits, we randomly sample 80\% of \textit{Waseem} as the train set, resulting in 8720 instances.
For \textit{Vidgen}, we use the original corpus-provided train split of 32924 instances. 
Since \textit{Ethos} has a relatively small size, we use the entire corpus for training, when used as the source. We call the source versions of these corpora as $\textit{Waseem}_\mathrm{src}$, $\textit{Vidgen}_\mathrm{src}$ and $\textit{Ethos}_\mathrm{src}$. Note that the source corpus is only used for training, while its validation set is not used for our experiments. Instead, we use the corresponding validation and test sets of the low-resource target corpus.

\paragraph{Target setting:} In order to simulate a low-resource scenario for the target, we down-sample the original training instances of the corpora 
to 500 instances. 
This yields three low-resource target corpora, namely, $\textit{Waseem}_\mathrm{tar}$, $\textit{Vidgen}_\mathrm{tar}$ and $\textit{Ethos}_\mathrm{tar}$. Furthermore, we split each of them in
the 80-20 ratio to obtain their respective low-resource train (400) and validation (100) sets. For the test set from $\textit{Waseem}_\mathrm{tar}$, we sample 10\% of the original data, disjoint from the train and validation sets, given by 1090 instances. We use the original test split of 4120 instances for $\textit{Vidgen}_\mathrm{tar}$.
For $\textit{Ethos}_\mathrm{tar}$, we randomly sample 20\% of the data, disjoint from the previous set of 500 instances, as the test set.

\subsection{Baselines}
We compare our approach with the following baseline approaches:

\paragraph{Target-FT:} We fine-tune the pre-trained BERT on the train set of the low-resource target corpus.
\paragraph{Seq-FT:} Here, we sequentially fine-tune the BERT model first on the resource-rich source corpus and then on the low-resource target corpus.
\paragraph{Mixed-FT:} Here, we fine-tune BERT on a mix of the source and target corpora. Since the target instances are limited, we first over-sample them. Then, for every mini-batch of size $m$, we randomly sample $m$ training instances each from the source and the target. We then combine their cross-entropy losses for updating the model parameters, as:
%This corresponds to Equation \ref{eq6}.
%\vspace{-0.6cm}
\begin{equation}\label{eq6}
\resizebox{0.84\columnwidth}{!}{%
$\underset{f,g}{\mathrm{min}} \quad \theta_s \frac{1}{m} \underset{i}{\sum} L_{s}(y_i^{s},f(g(x_i^{s}))) +  \theta_t \frac{1}{m} \underset{j}{\sum} L_{t}(y_j^{t}, f(g(x_j^t)))$
}
\end{equation}
This is similar to Equation \ref{eq3} without the OT$^{NN}$ losses.
\paragraph{\textit{k}NN-FT:} For every target instance, we retrieve top-$k$ neighbors from the source, ranked with cosine similarities over SBERT embeddings. This yields a subset of source instances that are neighbors to the target instances. We then fine-tune the BERT model with the strategy used for Mixed-FT.

\paragraph{\textit{k}NN ranking: } 
Here, we predict the labels of the target instances using a majority voting strategy. This voting is done over the labels associated with the top-\textit{k} retrieved neighbors from the source based on their cosine similarities.

\paragraph{Weighted \textit{k}NN: } This uses a weighted voting of the top-\textit{k} neighbors. Here we compute the sum of cosine similarities of neighbors associated with every class. The class with the highest score is returned as the predicted label of the target instance.

\paragraph{CE \textit{k}NN$^{+}$ + SRC:} This is the Cross-Encoder-based neighborhood framework \textit{k}NN$^{+}$, proposed by \citet{sarwar-etal-2022-neighborhood}, as discussed in Section \ref{related_neigh}.
For a fair comparison, we use the pre-trained BERT as the base representation. We first train CE \textit{k}NN$^{+}$ on the source (SRC) and then with the target instances and their neighbors from the source.

\paragraph{PretRand:} This is a transfer learning strategy proposed by \citet{Meftah2021NeuralSD} for low-resource domain adaptation. They jointly learn a pre-trained branch in the target model with a normalized, weighted, and randomly initialized branch. This is done so that the model can learn target-specific patterns while retaining the source knowledge. For a fair comparison, we use the pre-trained BERT as the base model,
which is first fine-tuned on the source. For the random branch, following the approach, we add a BiLSTM layer and a Fully Connected layer over the final hidden layer from BERT. The final predictions are obtained using an element-wise sum of the predictions from the two branches.

\paragraph{OT:} Finally, we use OT to transfer knowledge from the source to the target using both the ED and LC losses, similar to Equation \ref{eq3}. However, this is done \textit{without} incorporating any neighborhood information in both 
the 
cost matrix and the computation of $\gamma$. 

\begin{table*}[!t]
%\small
\scriptsize
\centering
%\resizebox{0.95\textwidth}{3.3cm}{%
\begin{tabularx}{0.85\textwidth}{p{0.6cm}|p{3.5cm}|p{1.4cm}|p{1.4cm}||p{1.4cm}|p{1.4cm}|| p{1.4cm}|p{1.4cm}}
\hline
\multicolumn{2}{l|}{\textbf{Target corpus}} &\multicolumn{2}{c||}{\textbf{Waseem$_{\mathbf{tar}}$}}&\multicolumn{2}{c||}{\textbf{Vidgen$_{\mathbf{\bm{tar}}}$}}&\multicolumn{2}{c}{\textbf{Ethos$_{\mathbf{\bm{tar}}}$}}  \\
\hline
\multicolumn{2}{l|}{Target-FT} & \multicolumn{2}{c||}{64.0$\pm$2.1} & \multicolumn{2}{c||}{68.8$\pm$3.2}  & \multicolumn{2}{c}{69.6$\pm$6.4} \\
\hline
\multicolumn{2}{l|}{\textbf{Source corpus}} & \textbf{ Vidgen$_{\mathbf{\bm{src}}}$} & \textbf{Ethos$_{\mathbf{\bm{src}}}$} & \textbf{ Waseem$_{\mathbf{\bm{src}}}$} & \textbf{ Ethos$_{\mathbf{\bm{src}}}$} & \textbf{Vidgen$_{\mathbf{\bm{src}}}$}&\textbf{ Waseem$_{\mathbf{\bm{src}}}$}   \\
\hline
\multicolumn{2}{l|}{Seq-FT} & 63.2$\pm$2.1 & 65.0$\pm$1.1 & 67.0$\pm$2.2 & 70.8$\pm$3.9 & \textbf{79.8}$\pm$0.7 & 70.2$\pm$3.1 \\
\multicolumn{2}{l|}{Mixed-FT} & 61.2$\pm$2.7 & 66.6$\pm$2.2 & 69.8$\mbox{*}\pm$1.6 & 71.4$\pm$3.9 & \underline{77.6}$\pm$2.1 & 71.8$\pm$3.5 \\
\multicolumn{2}{l|}{\textit{k}NN-FT} & 62.2$\pm$1.2 & 65.6$\pm$0.8 & 69.4$\mbox{*}\pm$2.3 & 70.8$\pm$1.9 & 77.2$\pm$1.5 & 70.6$\pm$3.4 \\
\multicolumn{2}{l|}{\textit{k}NN ranking} & 57.0 & 60.0 & 40.0 & 73.0$\mbox{*}$ & 77.0 & 49.0 \\
\multicolumn{2}{l|}{Weighted \textit{k}NN} & 57.0 & 60.0 & 37.0 & 73.0$\mbox{*}$ & 77.0 & 47.0  \\ 
\multicolumn{2}{l|}{CE \textit{k}NN${^{+}}$ + SRC} & 59.8$\pm$1.8 & \textbf{68.4}$\mbox{*}\pm$0.8 & 65.6$\pm$1.6 & 68.8$\pm$3.9 & 76.8$\pm$0.7 &  67.6$\pm$2.8 \\
\multicolumn{2}{l|}{PretRand} & 59.6$\pm$5.1 & 63.2$\pm$2.9 & \underline{71.0}$\mbox{*}\pm$0.6 & 72.2$\mbox{*}\pm$2.0 & \underline{77.6}$\pm$2.2 & 71.4$\pm$3.7 \\  \hline \hline
\multicolumn{2}{l|}{OT} & \underline{65.4}$\mbox{*}\pm$1.5 & 66.6$\pm$1.0 & 70.0$\mbox{*}\pm$2.8 & 71.4$\pm$5.2 & 73.6$\pm$3.6 & \textbf{74.6}$\mbox{*}\pm$2.9 \\ \hline
\multicolumn{2}{l|}{OT$^{NN}$} & \textbf{65.6}$\mbox{*}\pm$2.9 & \underline{67.4}$\mbox{*}\pm$1.6 & \textbf{71.6}$\mbox{*}\pm$1.4 & \underline{73.2}$\mbox{*}\pm$0.7 & 73.8$\pm$2.3 & 72.6$\mbox{*}\pm$3.1 \\
\multicolumn{2}{l|}{OT$^{NN}_{\mathrm{pre\mbox{-}select}}$} & 64.2$\pm$1.5 & 67.0$\pm$2.1 & \textbf{71.6}$\mbox{*}\pm$2.7 & 72.6$\mbox{*}\pm$1.0 & 75.4$\pm$1.4 & 73.2$\mbox{*}\pm$1.9 \\
\multicolumn{2}{l|}{OT$^{NN}$ + sloss} & 62.8$\pm$2.2 & \textbf{68.4}$\mbox{*}\pm$0.8 & 69.2$\mbox{*}\pm$3.2 & \textbf{73.8}$\mbox{*}\pm$1.6 & 76.8$\pm$1.9 & \underline{73.4}$\mbox{*}\pm$0.8 \\
\multicolumn{2}{l|}{OT$^{NN}_{\mathrm{pre\mbox{-}select}}$ + sloss} & 65.2$\mbox{*}\pm$1.7 & 66.6$\pm$1.6 & 70.2$\mbox{*}\pm$3.7 & 72.2$\mbox{*}\pm$1.3 & 77.2$\pm$1.3& \textbf{74.6}$\mbox{*}\pm$2.5  \\
%\bottomrule
\hline
\end{tabularx}
%}
\caption{\label{Results_OT}
F1 score ($\pm$std-dev) on the target corpus.  
The last four are the proposed OT$^{NN}$ variants. 
\textbf{Bold} denotes the best, \underline{underline} denotes the second-best scores in each column.  
$\mbox{*}$ denotes the significantly improved scores compared to Seq-FT using the McNemar test \citep{dror-etal-2018-hitchhikers,mcnemar1947note}.}
\end{table*}

\subsection{Hyper-parameters}
We train all the models for 10 epochs initialized with the pre-trained BERT-base \citep{devlin-etal-2019-bert} uncased model \citep{wolf-etal-2020-transformers}, with a maximum sequence length of 128 tokens. We use the Adam optimizer with a learning rate of $5 \times 10^{-5}$. Besides, we perform hyper-parameter tuning for $k$ and model selection using the best F1 scores over the respective target corpus validation sets. After the preliminary experiments, we set $\alpha$ = 0.05, $\beta$ = 10, $\epsilon$ = 0.2, $\lambda$ = 0.5,
and 
$\theta_t$ = 10  for all our experiments. 
We use a batch size of 32 for the OT$^{NN}$ and the baselines, except CE-\textit{k}NN$^{+}$. The latter inherently requires the batch size to be equal to the neighborhood size, as it provides query-neighborhood pairs as inputs to the model. See Appendix \ref{sec:impl_det} for further details on the hyper-parameter tuning.

\section{Results}

\subsection{Discussion}

Table \ref{Results_OT} shows the performance obtained with the baselines and the OT$^{NN}$ variants across the test sets of three low-resource target corpora using different resource-rich source corpora. We also present the performance with Target-FT for reference. Following the prior work on HS detection \citep{sarwar-etal-2022-neighborhood, attanasio-2022-entropy}, we use the F1 score of the hate class to report the performance, with an
average F1 computed over five runs of the same experiments with different random initializations.

The results show that transferring knowledge from a resource-rich corpus to a low-resource corpus is generally helpful. The best scores in the six respective settings of Table \ref{Results_OT} are substantially higher than those from Target-FT. Furthermore, while the baseline methods show inconsistent performance across different settings, the proposed OT$^{NN}$ variants yield the best performance in five out of six cases and the second-best in three cases. The baselines of Mixed-FT, $k$NN variants and CE $k$NN$^+$ achieve significant improvements compared to the vanilla Seq-FT for only 1 case, and PretRand achieves it for 2 cases. OT$^{NN}$ variants, on the other hand, yield significant improvements in most cases; for instance, OT$^{NN}$ has significantly improved scores in 5 out of 6 cases. Besides, the best scores from OT$^{NN}$ variants improve over OT in 5 settings, while staying on par with OT in the remaining setting. This demonstrates that incorporating neighborhood information results in a more effective transfer.

When $\textit{Vidgen}_\mathrm{src}$ is used for transferring knowledge to $\textit{Ethos}_\mathrm{tar}$, Seq-FT yields the highest score (79.8). This is apparently because $\textit{Vidgen}_\mathrm{src}$ comprises a wide range of hateful forms directed towards different social groups. Since $\textit{Ethos}_\mathrm{tar}$ also involves hate against a variety of social groups,  pre-training on all the source instances from $\textit{Vidgen}_\mathrm{src}$ for transfer learning, instead of training with the nearest neighbors, seems to be more helpful in this case.
However, this is not the case when the transfer occurs from $\textit{Ethos}_\mathrm{src}$ to $\textit{Vidgen}_\mathrm{tar}$. This is likely because the \textit{Vidgen} corpus involves adversarial instances that can easily fool an HS detection system trained on a different corpus. Besides, $\textit{Ethos}_\mathrm{src}$ has a subset of hateful forms and social groups covered by \textit{Vidgen}. Therefore, a nearest neighborhood framework %from $\textit{Vidgen}_\mathrm{tar}$ 
for transferring knowledge from  $\textit{Ethos}_\mathrm{src}$ to $\textit{Vidgen}_\mathrm{tar}$ 
yields an improved performance, the highest score being 73.8 obtained by OT$^{NN}$ + sloss, compared to 70.8 from Seq-FT.

\begin{figure}

%\adjustbox{max width=\columnwidth}{% 
\resizebox{0.49\columnwidth}{3.4cm}{%
\begin{tikzpicture}
 
\begin{axis} [ybar = .05cm,
    bar width = 8pt,
    ymin = 60, 
    ymax = 80, 
    enlarge y limits = {abs = .8, upper},
    enlarge x limits = {abs = 0.5},
     legend style={nodes={scale=0.95, transform shape}, at={(0.03,0.85)},anchor=west},
     xtick={1,2,3,4},
     xticklabels={240 (300),400 (500),560 (700),720 (900)},
     xlabel=(a) \textit{Ethos}${_{\mathrm{src}}}$ to \textit{Vidgen}${_{\mathrm{tar}}}$,
   %  ticklabel style = {font=\normal},
     label style={font=\LARGE},
     ylabel=F1 score,
]
 
\addplot coordinates {(1,67) (2, 68.8) (3, 66.6) (4,70)};
\addplot coordinates {(1, 62.8) (2, 71.4) (3, 66.4) (4,72)};
\addplot coordinates {(1, 70) (2,73.8) (3,72.4) (4,73.2)};

\legend{Target-FT, Mixed-FT, OT$^{NN}$ + sloss}
 
\end{axis}
\end{tikzpicture}
}
\resizebox{0.49\columnwidth}{3.4cm}{%
\begin{tikzpicture}

\begin{axis} [ybar = .05cm,
    bar width = 12pt,
    ymin = 60, 
    ymax = 80, 
    enlarge y limits = {abs = 1.0, upper},
    enlarge x limits = {abs = .5},
    legend style={nodes={scale=0.9, transform shape}},
    xtick={1,2,3},
    label style={font=\LARGE},
    xticklabels={240 (300),400 (500),560 (700)},
    xlabel=(b) \textit{Waseem}${_{\mathrm{src}}}$ to \textit{Ethos}${_{\mathrm{tar}}}$, 
    ylabel=F1 score,
]
 
\addplot coordinates {(1,65.2) (2, 69.6) (3, 71.8)};
\addplot coordinates {(1, 69.6) (2, 71.8) (3, 70)};
\addplot coordinates {(1, 75.8) (2, 74.6) (3, 72.4)};

 \legend{Target-FT, Mixed-FT, OT$^{NN}_\mathrm{pre\mbox{-}select}$ + sloss}
 
\end{axis}

\end{tikzpicture}
}
\caption{\label{fig1} Performance with different sizes of the target train set.  
The total number of labeled instances available from the target is mentioned within the brackets, where the remaining instances are used as the target validation set.}
\end{figure}
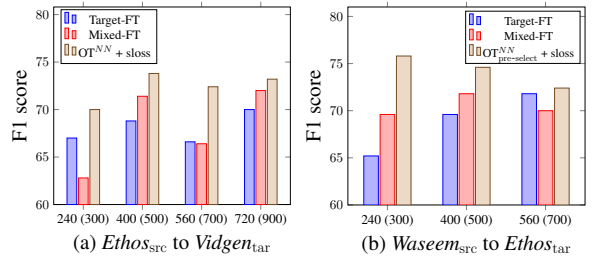

\begin{table*}[!t]
%\small
\scriptsize
\centering
\begin{tabularx}{0.92\textwidth}{p{0.6cm}|p{3.5cm}|p{1.4cm}|p{1.4cm}||p{1.4cm}|p{1.4cm}|| p{1.4cm}|p{1.4cm}}
\hline
\multicolumn{2}{l|}{\textbf{Target corpus}} &\multicolumn{2}{c||}{\textbf{Waseem$_{\mathrm{\bm{tar}}}$}}&\multicolumn{2}{c||}{\textbf{Vidgen$_{\mathrm{\bm{tar}}}$}}&\multicolumn{2}{c}{\textbf{Ethos$_{\mathrm{\bm{tar}}}$}}  \\
\hline
\multicolumn{2}{l|}{\textbf{Source corpus}} & \textbf{ Vidgen$_{{\mathbf{\bm{src}}}}$} & \textbf{Ethos$_{\mathbf{\bm{src}}}$} & \textbf{ Waseem$_{\mathbf{\bm{src}}}$} & \textbf{ Ethos$_{\mathbf{\bm{src}}}$} & \textbf{ Vidgen$_{\mathbf{\bm{src}}}$}&\textbf{ Waseem$_{\mathbf{\bm{src}}}$}   \\
\hline
\multicolumn{2}{l|}{OT$^{NN}$ + sloss} & \underline{62.8}$\pm$2.2 & \textbf{68.4}$\pm$0.8 & \textbf{69.2}$\pm$3.2 & \textbf{73.8}$\pm$1.6 & \textbf{76.8}$\pm$1.9 & \textbf{73.4}$\pm$0.8 \\
\multicolumn{2}{l|}{OT$^{NN}$ + sloss (without ED)} & \textbf{63.8}$\pm$1.3 & 65.8$\pm$1.7 & \underline{68.0}$\pm$0.0 & 70.0$\pm$2.4 & \underline{76.4}$\pm$0.8  & \underline{71.8}$\pm$2.5 \\
\multicolumn{2}{l|}{OT$^{NN}$ + sloss (without LC)} & 62.0$\pm$2.1 & \underline{66.4}$\pm$2.2 & 67.6$\pm$2.7 & \underline{72.4}$\pm$1.4 & 75.2$\pm$2.6 & 67.8$\pm$3.9 \\ \hline \hline
\multicolumn{2}{l|}{OT$^{NN}_{\mathrm{pre\mbox{-}select}}$ + sloss} &\textbf{65.2}$\pm$1.7 & \underline{66.6}$\pm$1.6 & \textbf{70.2}$\pm$3.7 & \textbf{72.2}$\pm$1.3 & \textbf{77.2}$\pm$1.3& \textbf{74.6}$\pm$2.5  \\
\multicolumn{2}{l|}{OT$^{NN}_{\mathrm{pre\mbox{-}select}}$ + sloss (without ED)} & \underline{64.4}$\pm$1.5 & \textbf{67.6}$\pm$1.4 & \underline{67.6}$\pm$4.3 & 70.8$\pm$2.3 & \underline{75.6}$\pm$2.7 & \underline{74.2}$\pm$5.6 \\
\multicolumn{2}{l|}{OT$^{NN}_{\mathrm{pre\mbox{-}select}}$ + sloss (without LC)} & 62.2$\pm$2.6 & 63.8$\pm$1.5 & 67.2$\pm$5.0 & \underline{71.8}$\pm$1.5 & 74.6$\pm$4.1 & 67.2$\pm$5.2  \\
%\bottomrule
\hline
\end{tabularx}
%}
\caption{\label{Loss_ablation}
Ablation study without the Embedding Distance (ED) \slash Label Consistency (LC) losses. F1 ($\pm$std-dev) on low-resource target corpus. 
\textbf{Bold} denotes the best, \underline{underline} denotes the second-best score for each OT${^{NN}}$ variant.}
\end{table*}

\paragraph{Varying the size of $\bm{X^t}$:} We vary the size of the labeled target corpus available for training. We illustrate the cases of transferring knowledge from  $\textit{Ethos}_\mathrm{src}$ to $\textit{Vidgen}_\mathrm{tar}$ in Figure \ref{fig1}(a), and from $\textit{Waseem}_\mathrm{src}$ to $\textit{Ethos}_\mathrm{tar}$ in Figure \ref{fig1}(b), with different OT$^{NN}$ variants. For $\textit{Vidgen}_\mathrm{tar}$, we sample 300, 500, 700, and 900 instances. We use 80\% for training, given by 240, 400, 560, and 720 instances, respectively, and the remaining 20\% for validation. Since the $Ethos$ corpus is small, we sample only 300, 500, and 700 instances as $\textit{Ethos}_\mathrm{tar}$, with the same proportions for training and validation. The target test set remains the same as in Table \ref{dataset-prop} for different training sizes. 
We observe that the OT$^{NN}$ variants consistently improve the performance, with larger improvements obtained when the size of available target instances is lower. Mixed-FT, on the other hand, is inconsistent, and in some cases performs worse than Target-FT.

The improvements with OT$^{NN}$ can be attributed to 
the fact that it can  
systematically \textit{learn} the amount of transfer 
based on both the embedding distance and label consistency.

\begin{table*}[!h]
\scriptsize
%\centering
\begin{center}
\begin{tabularx}{\textwidth}{ p{14.5cm} | p{1.1cm} }
\hline
\multicolumn{2}{c}{\textbf{Target instance:} rt i am not sexist but girls who smoke are a big turn off even if your supermodel status its still a turn off. }\\
\hline
\multicolumn{1}{c|}{\bf Instances}  & \multicolumn{1}{l}{\bf GT}   \\ 
\hline
\multicolumn{2}{c}{\textbf{Nearest neighbors from source using SBERT}} \\
\hline
i do not actually care much for looks but the inflated sense of self importance that girls seem to have nowadays is rather obnoxious. & hate \\ 
totally agree. i could never find a woman or man attractive to go out with.
....
i could nt be in a relationship with them as i am
 aromantic. & non-hate \\
every guy i know at work smoke at least 10 fags a day and i call them out. do not like it then do not do it. & non-hate \\
wow the truth is i did not get turned on until you mentioned cigar smoking men pissing on other men & non-hate \\
 women have no redeeming features & hate \\
 i am not sexist but women are dumb & hate \\
 weed is stoner trash. if you are not a stoner guy, smoking this estrogen-filled garbage makes you a honorary one. & non-hate\\
 maybe i am sexist because women are dumb & hate \\
 women are less attracted to looks. they are far more attracted to high status and a dominant personality. the good news is you can acquire both of these. & non-hate\\
 women without big t\mbox{*}ts are not attractive & hate\\
\hline
\multicolumn{2}{c}{\textbf{Nearest neighbors from source using OT$^{\bm{NN}}$}} \\
\hline
 i am not sexist but women are dumb & hate\\
 maybe i am sexist because women are dumb & hate\\
 thrilled to have been born as the superior male gender & hate\\
 men are superior than women in 
every single aspect & hate \\
women of color are not pleasant to look at & hate \\
leadership positions should only be occupied by men. that is 
the way to get things done. & hate \\
all women are created as equal b$\mbox{*}$tches & hate \\
all women should be suppressed & hate \\
this is the fact feminism is girlf$\mbox{*}$rts you cant unsmell & hate \\
women are devious & hate \\
\bottomrule
\end{tabularx}
\end{center}
%\textbf{\textbf{}}}
\caption{\label{quali-analysis} Qualitative analysis of an example with its top 10 nearest neighbors extracted using the SBERT and the learned OT$^{NN}$ representations, where the source is \textit{Vidgen}$_\text{src}$ and the target is \textit{Waseem}$_\text{tar}$; GT: Ground Truth class.}
\end{table*}

\begin{figure*}
\resizebox{\textwidth}{5.5cm}{%
\begin{tikzpicture}
%\resizebox{\textwidth}{3.4cm}{%
    \begin{groupplot}[group style={group size= 3 by 1, horizontal sep=2cm},height=5.5cm,width=\textwidth]
        \nextgroupplot[xlabel=$\#$neighbors,
	ylabel=F1 score,
	width=10cm,height=7cm,
	label style={font=\huge},
	legend style={font=\LARGE},
	title style={font=\huge},
	ticklabel style = {font=\Large},
	title={$\textit{Ethos}_\text{src}$ to $\textit{Vidgen}_\text{tar}$},
	line width=3pt,legend to name=zelda]
    \addplot[color=red,mark=x, mark size=5pt] coordinates {
	(10, 69)
	(30, 73)
	(50, 73)
	(70, 73)
	(100, 73)
};\addlegendentry{SBERT ${k}$NN ranking};
                \addplot[color=blue,mark=*, mark size=3pt] coordinates {
	(10, 73.07)
 	(30, 73.28 )
	(50, 73.22)
	(70, 73.19)
	(100,73.21)
};\addlegendentry{OT$^{NN}$ ${k}$NN ranking};
        \nextgroupplot[xlabel=$\#$neighbors,
	ylabel=F1 score,
	width=10cm,height=7cm,
	label style={font=\huge},
	title style={font=\huge},
	ticklabel style = {font=\Large},
	title={$\textit{Waseem}_\text{src}$ to $\textit{Ethos}_\text{tar}$},
	line width=3pt]
                \addplot[color=red,mark=x,mark size=5pt]coordinates {
	(10, 46)
	(30, 48)
	(50, 52)
 	(70, 49)
	(100, 51)
};
\addplot[color=blue,mark=*, mark size=3pt] coordinates {
	(10, 73.80)
	(30, 73.68)
	(50, 74.47)
	(70, 74.47)
	(100, 73.80)
};

 \nextgroupplot[xlabel=$\#$neighbors,
	ylabel=F1 score,
	width=10cm,height=7cm,
	label style={font=\huge},
	title style={font=\huge},
	ticklabel style = {font=\Large},
    title={$\textit{Vidgen}_\text{src}$ to $\textit{Waseem}_\text{tar}$},
    line width=3pt]
    \addplot[color=red,mark=x, mark size=5pt] coordinates {
	(10, 48)
	(30, 52)
	(50, 55)
	(70, 57)
	(100, 57)
};

\addplot[color=blue,mark=*, mark size=3pt] coordinates {
	(10, 63.07)
	(30, 64.10)
	(50, 63.34)
	(70, 63.48)
	(100, 63.29)
};
    \end{groupplot}

    \node[below,inner sep=55pt] at(current bounding box.south) {\pgfplotslegendfromname{zelda}};
\end{tikzpicture}
}
\caption{
F1 using the majority voting of the $k$-Nearest Neighbors retrieved from SBERT and OT$^{NN}$ representations.}
\label{kNN_rank}
\end{figure*}
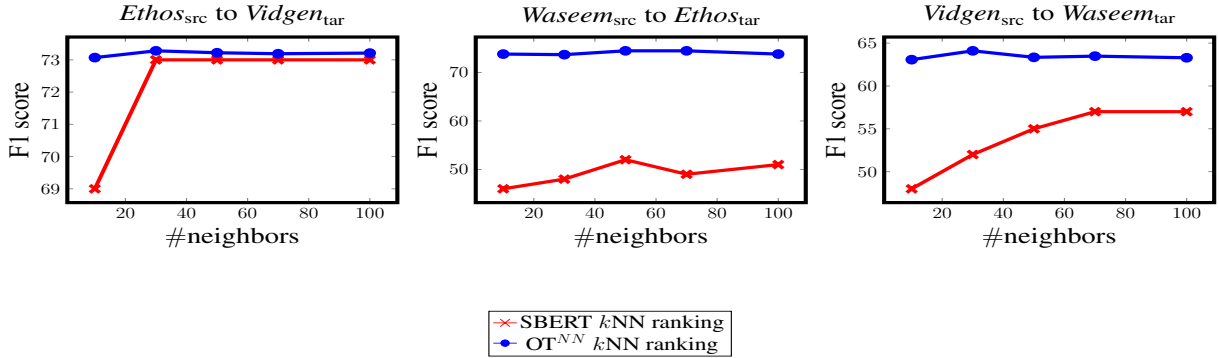

\subsection{Ablation Study}

To analyze the importance of the constituent losses in OT$^{NN}$, we present an ablation study in Table \ref{Loss_ablation} by removing the ED and the LC losses from the OT$^{NN}$ variants. 
The best performances for each variant are obtained in 5 out of 6 cases when both the ED and the LC losses are incorporated.
Besides, the second-best performances are obtained, in most cases, when we remove the ED loss. This suggests that while both losses are essential for an effective transfer, the LC loss contributes more towards the final performance than the ED loss.  

\subsection{Analysis of OT$^{\bm{NN}}$ Representations}

We analyze the effect of training with OT$^{NN}$ on the representation space by extracting the nearest neighbors of target instances. We rank these neighbors with cosine similarity over the learned OT$^{NN}$ representations and check their ground truth classes. We compare them with the nearest neighbors obtained using SBERT representations. Table \ref{quali-analysis} contains an example of a hateful instance from \textit{Waseem}$_\text{tar}$, and its top 10 nearest neighbors from \textit{Vidgen}$_\text{src}$.
We observe that the neighbors retrieved using the SBERT representations belong to both hate and non-hate classes. This is because SBERT is optimized mainly for semantic similarity, while they are sub-optimal in differentiating hateful instances from non-hateful ones. On the other hand, the neighbors obtained from OT$^{NN}$ representations 
indicate that OT$^{NN}$ brings instances across corpora, which are both semantically similar (the topic of women) and belong to the same class closer in the representation space, compared to those belonging to the opposite class.

In addition, we study the effect of the OT$^{NN}$
representations by performing a simple majority
voting of the top $k$ nearest neighbors retrieved from the source with SBERT versus OT$^{NN}$. Figure \ref{kNN_rank} demonstrates the performance obtained on the target test set.
Here the neighbors from the two representation spaces are ranked using cosine similarities. We can see that majority voting using 
the OT$^{NN}$ representations achieves higher performance compared to that using the SBERT representations for different numbers of neighbors.

\section{Conclusion and Future Work}
In this work, we proposed a framework for transferring knowledge to a low-resource HS corpus by incorporating neighborhood information with Optimal Transport. It 
allowed the model to flexibly learn the amount of transfer from the nearest neighbors based both on their proximity in a sentence embedding space and label consistency. Our framework yielded substantial improvements across HS corpora from varied platforms in low-resource settings. 
Besides, the qualitative analysis of its learned representations
demonstrated that they incorporate both semantic 
and label similarities. This is different from sentence embedding representations, where semantically similar instances may have opposite labels. 

Since our framework uses neighborhood information for transferring knowledge, it relies on the degree of proximity of the neighbors. However, if all of the source and target instances are very distant semantically, all the nearest neighbors from the source may have very low cosine similarity to the corresponding target instances. In such scenarios, the framework may yield limited improvements over the vanilla fine-tuning as the available neighborhood information would be much weaker. In such cases, the performance would mainly depend on the label consistency of the neighbors.

For future work, 
our framework can be explored for transferring knowledge from resource-rich languages, 
such as English, 
to low-resource languages. This can be done by extracting the cross-lingual neighbors using multilingual sentence embedding models like LaBSE \citep{feng-etal-2022-language}.
Besides, the framework can be applied for transferring knowledge in other text classification tasks, such as sentiment classification, bragging detection \citep{jin-etal-2022-automatic}, etc., as the methodology is not restricted to only hate speech detection.

\section*{Ethical Considerations}
The proposed approach intends to support more robust detection of online hate speech that can use the existing annotated resources for transferring knowledge to a resource with limited annotations. We acknowledge that annotating hateful content can have negative effects on the mental health of the annotators. The corpora used in this work are publicly available and cited appropriately in this paper. The authors of the respective corpora have provided detailed information about the sampling strategies, data collection process, annotation guidelines, and annotation procedure in peer-reviewed articles. Besides, the hateful terms and slurs presented in the work are only intended to give better insights into the models for research purposes. 

% While identifying individual hateful instances in social media can aid manual content moderators, completely preventing the occurrence and spread of online hate speech is more challenging. This is because hate speech is a result of deeper social stereotypes and coordinated attacks from communities with vested interests. Different hateful forms evolve with time on online platforms, which are a result of the ever-changing world politics and social structure. This requires that a broader knowledge about the stereotypes existing in society be incorporated into the machine-leaning models so that they can identify them. This is because such stereotypes facilitate the origin and spread of hate.   

\section*{Acknowledgements}
This work was supported partly by the french PIA project ``Lorraine Université d'Excellence'', reference ANR-15-IDEX-04-LUE. Experiments presented in this article were carried out using the
Grid'5000 testbed, supported by a scientific interest group hosted by Inria and including CNRS, RENATER and several Universities as well as other organizations (see \url{https://www.grid5000.fr}). We are extremely grateful to Claire Gardent for taking time out to review the paper internally and Michel Olvera for his very helpful feedbacks regarding the work and for internally reviewing the paper. We would also like to thank the anonymous reviewers for their valuable feedbacks and suggestions.

\bibliography{anthology,acl}
\bibliographystyle{acl_natbib}

\appendix
\section{Illustration of OT$^{\bm{NN}}$}
\label{OT-fig}

\begin{figure*}[t]
\begin{minipage}{\textwidth}
\centering 
\includegraphics[width=1.05\textwidth, height=7.1cm]{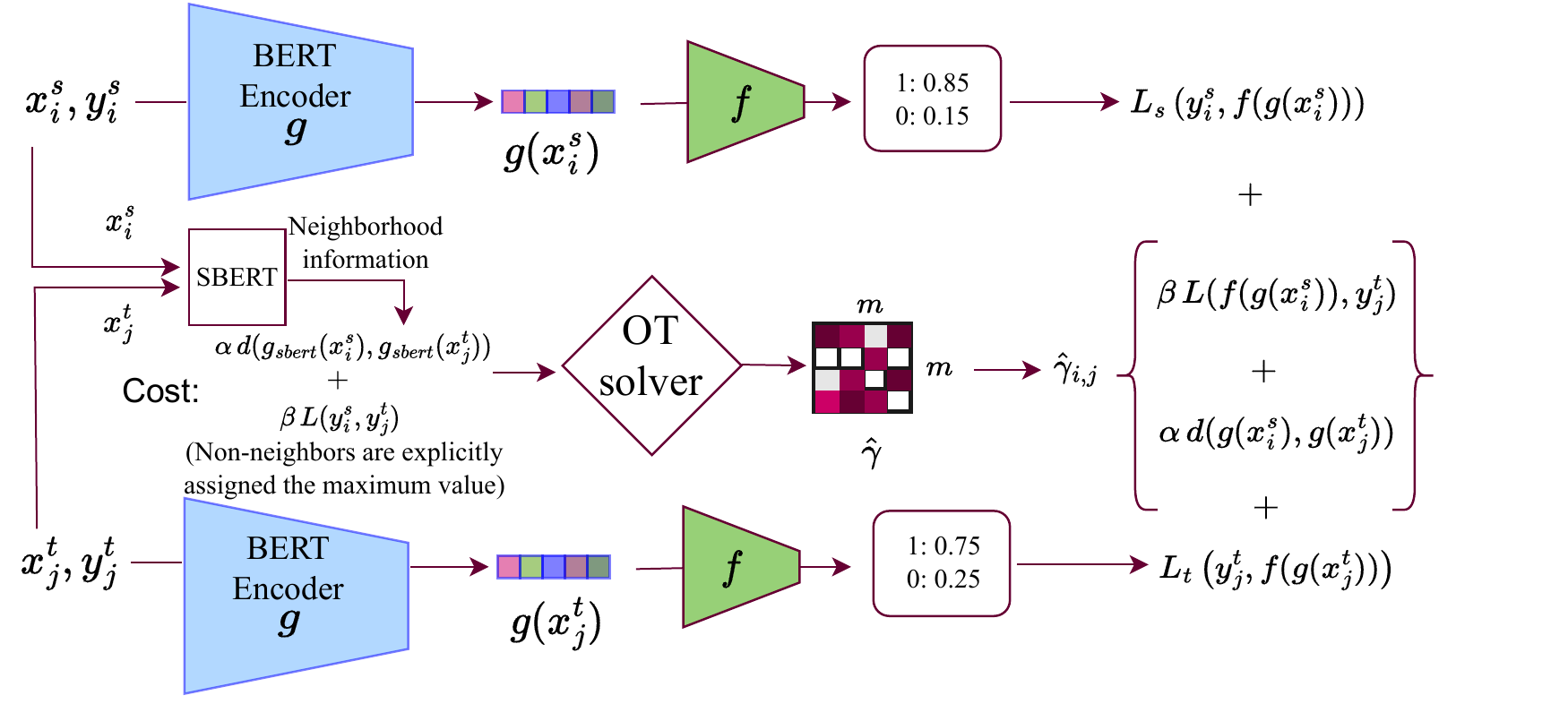} \qquad
\captionof{figure}{Illustration of the training strategy in OT$^{NN}$. Even though the BERT encoder $g$ and the classifier $f$ are shared by both corpora, they are illustrated
twice for better clarity by representing the two corpora separately. The presented softmax values obtained from $f$ are simply examples provided for illustration. The figure is inspired by \citet{damodaran2018deepjdot}.} 
\label{OT_NN_fig}
\end{minipage}
\end{figure*}

Figure \ref{OT_NN_fig} presents an illustration of the proposed $OT^{NN}$ training strategy.

\section{Corpus Details}
\label{corp_det}

The corpora used in our experiments are collected during different time periods, with different sampling strategies across varied online platforms. Following are some additional details about the corpora discussed in Section \ref{corp}.

\textbf{Waseem:} This Twitter corpus, provided by \citet{waseem}, is sampled mainly using keywords containing common terms and slurs associated with
hate against sexual, gender, religious, and ethnic minorities. 
It originally has three classes: `sexism', `racism', and `none'. However, 
since it is made available as tweet IDs, we observe that a major portion of the racist comments is removed by Twitter, and we could retrieve only 20 instances labeled as `racism'. 
Due to the difference in the labeling schemes across corpora, we follow other prior work for cross-corpus evaluations \citep{wiegand-etal-2019-detection, swamy-etal-2019-studying}, 
and merge the original `racism' and `sexism' classes into the class \textit{hate}. In total, we obtain 10.9K tweets, of which 26.8\% instances belong to the \textit{hate} class. A major portion of hateful comments in this corpus is subtle and do not involve profane words \citep{wiegand-etal-2019-detection}. 

\textbf{Vidgen:} The \textit{Vidgen} corpus \citep{vidgen-etal-2021-learning} is dynamically generated using a human-and-model-in-the-loop process that is aimed at making the corpus robust to different forms of hate. These hateful forms are derogation, animosity, threatening language, support for hateful entities, and dehumanization. It covers hate directed against diverse social groups like blacks, women, jews, muslims, immigrants, arabs, disabled people, trans people, and others. It has adversarial content that can mislead the HS detection systems trained on other corpora, such as `\textit{all women are created equal, except chinks}', `\textit{the tea cup is bigger than the f*cking teapot}'. We use its version v0.2.3 containing a total of 41144 instances, where duplicates are removed, of which 53.9\% is labeled as \textit{hate}.

\textbf{Ethos:} This corpus \citep{mollas_ethos_2022} consists of 998 comments from YouTube and Reddit, sampled using an active learning strategy, which ensures both diversity and balance with respect to different hateful aspects defined. One of the ways they ensure this is by keeping only one instance of multiple comments with similar meanings. For example, the comments `\textit{I hate white people}' and `\textit{I hate whites}' \citep{mollas_ethos_2022} are similar, and only one of them is added.
It includes hate directed towards diverse identities, such as gender, race, national origin, disability, religion, and sexual orientation. In this work, we use the binary version of this corpus with 43.4\% \textit{hate} instances. 

\begin{table*}[!t]
%\small
\scriptsize
\centering
%\resizebox{0.95\textwidth}{3.3cm}{%
\begin{tabularx}{0.8\textwidth}{p{0.6cm}|p{3.5cm}|p{1.4cm}|p{1.4cm}||p{1.4cm}|p{1.4cm}|| p{1.4cm}|p{1.4cm}}
\toprule
\multicolumn{2}{l|}{\textbf{Target corpus}} &\multicolumn{2}{c||}{\textbf{Waseem$_{\mathbf{\bm{tar}}}$}}&\multicolumn{2}{c||}{\textbf{Vidgen$_{\mathbf{\bm{tar}}}$}}&\multicolumn{2}{c}{\textbf{Ethos$_{\mathbf{\bm{tar}}}$}}  \\
\hline
\multicolumn{2}{l|}{\textbf{Source corpus}} & \textbf{ Vidgen$_{\mathbf{\bm{src}}}$} & \textbf{Ethos$_{\mathbf{\bm{src}}}$} & \textbf{ Waseem$_{\mathbf{\bm{src}}}$} & \textbf{ Ethos$_{\mathbf{\bm{src}}}$} & \textbf{ Vidgen$_\mathbf{\bm{src}}$}&\textbf{ Waseem$_\mathbf{\bm{src}}$}   \\
\hline
\multicolumn{2}{l|}{Seq-FT} & 63.2$\pm$2.1 & 65.0$\pm$1.1 & 67.0$\pm$2.2 & 70.8$\pm$3.9 & 79.8$\pm$0.7 & 70.2$\pm$3.1 \\ \hline
\multicolumn{2}{l|}{$k$ = 10}& 59.8$\pm$1.8 & 68.4$\pm$0.8 & 65.6$\pm$1.6 & 68.8$\pm$3.9 & 76.8$\pm$0.7 &  67.6$\pm$2.8 \\
\multicolumn{2}{l|}{$k$ = 20} & 61.2$\pm$1.5 & 67.6$\pm$1.5 & 64.8$\pm$1.6 & 69.2$\pm$3.2 & 76.8$\pm$1.0  & 67.4$\pm$3.3 \\ 
\multicolumn{2}{l|}{$k$ = 30} & 60.3$\pm$1.6 & 68.1$\pm$1.0 & 64.4$\pm$1.9 & 69.9$\pm$2.8  & 76.8$\pm$0.5 & 68.5$\pm$1.7 \\
\multicolumn{2}{l|}{$k$ = 40} & 61.6$\pm$1.6 & 68.6$\pm$1.4 & 64.6$\pm$1.0 & 70.8$\pm$3.5 & 76.2$\pm$1.2 & 68.2$\pm$2.6 \\
\multicolumn{2}{l|}{$k$ = 50} & 60.8$\pm$2.0 & 68.8$\pm$0.7 & 62.8$\pm$2.6 & 68.4$\pm$4.8 & 75.8$\pm$0.4 & 68.4$\pm$0.5\\

\hline
\end{tabularx}
%}
\caption{\label{CE_kNN}
%\normalsize
Performance of CE $k$NN$^+$ + SRC with different neighborhood sizes, compared with Seq-FT. F1 score ($\pm$std-dev) is reported on the low-resource target corpus with 400 labeled training instances (total 500 labeled instances from the target) available.}
\end{table*}

\section{Data Preprocessing}
\label{pre-proc}
We pre-process the corpora by removing the URLs, splitting the hashtags into constituent words using CrazyTokenizer\footnote{\url{https://redditscore.readthedocs.io}}, expanding contractions (e.g. i'll to i will), and removing the rarely occurring Twitter handles and numbers. We finally convert the instances into lower case.

\section{Implementation Details}
\label{sec:impl_det}
For implementing the proposed OT$^{NN}$ framework, we fine-tune the pre-trained BERT-base uncased model, implemented by Hugging Face \citep{wolf-etal-2020-transformers},
having 110 million parameters, with the joint distribution OT framework\footnote{\url{https://github.com/bbdamodaran/deepJDOT}}. We encode an instance into the embedding space by obtaining the representations of the [CLS] token from the last hidden layer of BERT, which is a 768-dimensional vector in the BERT-base. We fine-tune the BERT model end-to-end for the classification task. Therefore, the [CLS] representations are the fine-tuned BERT representations. For incorporating the neighborhood information, we use the pre-trained SBERT sentence embeddings from `all-mpnet-base-v2'\footnote{\url{https://huggingface.co/sentence-transformers/all-mpnet-base-v2}} model, which is a sentence transformer model. 
For computing $\gamma$, we use the entropic regularized unbalanced OT solver using the Python Optimal Transport package\footnote{\url{https://pythonot.github.io/gen_modules/ot.unbalanced.html\#ot.unbalanced.sinkhorn_unbalanced}} \citep{JMLR:v22:20-451} at the mini-batch level. 

For the baselines of ${k}$NN-FT, $k$NN ranking, weighted $k$NN  and the OT$^{NN}$ variants, we select the number of neighbors ($k$)  from the range \{10, 30, 50, 70, 100, 200, 300, 400, 500\} through tuning over the corresponding target validation sets with respect to the F1 score of the hate class with a random seed. We set $\alpha$ = 0.05 and $\beta$ = 10 in Equation \ref{eq2} and \ref{eq4}, and $\theta_s$ = 1 for OT$^{NN}$ / $\mathrm{OT}_\mathrm{pre-select}^{NN}$ + sloss and $\theta_t$ = 10 in Equation \ref{eq3}, \ref{eq5} and \ref{eq6} for all the experiments. For OT$^{NN}$ without sloss, we set $\theta_s$ = 0.

For CE $k$NN$^+$ + SRC, we perform experiments with the implementation provided to us by the authors and report the results for the neighborhood size of 10 in Table \ref{Results_OT}. Even though \citet{sarwar-etal-2022-neighborhood} use 10 as the neighborhood size in their task of transfer learning in a cross-lingual set-up, we experiment with different neighborhood sizes ($k$ values). The results are reported in Table \ref{CE_kNN}. However, we could not increase the neighborhood size beyond 50 because of resource constraints. This is because a mini-batch in their framework comprises a query instance from the target and all its $k$ neighbors from the source. Thus, the number of neighbors is limited by the mini-batch size, which usually needs to be kept small when fine-tuning large language models like BERT. We can observe from Table \ref{CE_kNN} that the performances obtained with different neighborhood sizes are similar.

% \citet{sarwar-etal-2022-neighborhood} apply their CE $k$NN$^+$ + SRC approach for cross-lingual content flagging including toxicity detection. Toxicity is a
% broader concept compared to
% %are more compared to
% %conceptually different from 
% hate speech. This is because the mere presence of profane words can also make a comment toxic but not necessarily hateful \citep{Malmasi2018ChallengesID}. Hateful comments 
% %can be more subtle and 
% typically require some distinct targets, e.g. social groups (especially minorities) against whom the hateful content is directed, with or without using profane words \citep{waseem}, which makes hate speech detection a more challenging task \citep{Malmasi2018ChallengesID}.

\begin{table*}[!t]
\scriptsize
\centering
%\resizebox{0.95\textwidth}{3.3cm}{%
\begin{tabularx}{0.85\textwidth}{p{0.6cm}|p{3.5cm}|p{1.4cm}|p{1.4cm}||p{1.4cm}|p{1.4cm}|| p{1.4cm}|p{1.4cm}}
\toprule
\multicolumn{2}{l|}{\textbf{Target corpus}} &\multicolumn{2}{c||}{\textbf{Waseem$_\mathbf{\bm{tar}}$}}&\multicolumn{2}{c||}{\textbf{Vidgen$_\mathbf{\bm{tar}}$}}&\multicolumn{2}{c}{\textbf{Ethos$_\mathbf{\bm{tar}}$}}  \\
\hline
\multicolumn{2}{l|}{\textbf{Source corpus}} & \textbf{ Vidgen$_\mathbf{\bm{src}}$} & \textbf{Ethos$_\mathbf{\bm{src}}$} & \textbf{ Waseem$_\mathbf{\bm{src}}$} & \textbf{ Ethos$_\mathbf{\bm{src}}$} & \textbf{ Vidgen$_\mathbf{\bm{src}}$}&\textbf{ Waseem$_\mathbf{\bm{src}}$}   \\
\hline
\multicolumn{2}{l|}{Mixed-FT} & 17.8 m & 0.4 m & 4.7 m & 0.5 m & 14.0 m & 4.7 m\\ \hline
\multicolumn{2}{l|}{OT$^{NN}$} & 18.9 m & 0.4 m & 5.1 m & 0.6 m & 14.2 m & 5.0 m  \\
\multicolumn{2}{l|}{OT$^{NN}_\mathrm{pre\mbox{-}select}$} & 3.7 m  & 0.3 m  & 1.1 m & 0.6 m & 6.5 m & 3.4 m \\
\multicolumn{2}{l|}{OT$^{NN}$ + sloss} & 18.9 m & 0.4 m & 5.0 m & 0.6 m & 14.5 m & 4.9 m \\
\multicolumn{2}{l|}{OT$^{NN}_\mathrm{pre\mbox{-}select}$ + sloss} & 11.7 m & 0.4 m & 3.8 m & 0.6 m & 5.5 m & 3.9 m  \\

\hline
\end{tabularx}
%}
\caption{\label{Computation_time}
Per epoch training time in minutes for different settings.}
\end{table*}

We implement PretRand ourselves following the description provided by \citet{Meftah2021NeuralSD}. This approach is evaluated by the authors on the tasks of part-of-speech tagging, chunking, named entity recognition, and morphosyntactic tagging. Therefore, the approach uses a sequence labeling model with pre-trained word embeddings and a BiLSTM-based feature extractor. However, for a fair comparison with our approach, we use the pre-trained BERT model as the feature extractor instead of the BiLSTM model for the pre-trained units. For the randomly initialized units, we follow the approach and add a BiLSTM layer over the last hidden layer of the BERT model. We first fine-tune the pre-trained BERT model, without the randomly initialized units, on the source corpus. We then fine-tune the model with the additional randomly initialized units on the target corpus. We use the Adam optimizer with a learning rate of $5 \times 10^{-5}$ for the pre-trained BERT parameters. For the randomly initialized units, we use the Adam optimizer with a learning rate of $1.5 \times 10^{-2}$ following \citet{Meftah2021NeuralSD}.

\section{Computational Efficiency}
\label{sec:comp}

We present the per epoch training time of Mixed-FT and OT${^{NN}}$ variants for different settings of the source and target corpora in Table \ref{Computation_time}. Mixed-FT is a baseline that involves training the pre-trained BERT model on the combination of the source and target corpora. For every mini-batch of size $m$, there are $m$ instances sampled from each of the source and target corpora (Equation \ref{eq6}). This is the same mini-batch sampling that is followed in OT${^{NN}}$. We use one Nvidia GTX 1080 Ti GPU for our experiments. We can observe that OT${^{NN}}$ results in approximately the same computation time as taken by Mixed-FT in most of the settings as it does not change the model architecture, but only introduces a new training strategy. With the `pre-select' variant, the computation time gets further reduced in a few settings. This is because, in this variant, the model only gets trained on a subset of pre-selected source instances based on the neighborhood size.

\end{document}